\documentclass[10pt,twocolumn,letterpaper]{article}
\usepackage[pagenumbers]{cvpr} % To force page numbers, e.g. for an arXiv version

\usepackage{graphicx}
\usepackage{amsmath,bm}
\usepackage{amssymb}
\usepackage{booktabs}
\usepackage{array}
\usepackage{comment}
\usepackage[accsupp]{axessibility}
\newcommand{\argmax}{\mathop{\mathrm{argmax}}}

\newcolumntype{x}[1]{>{\centering\arraybackslash\hspace{0pt}}p{#1}}
\usepackage{caption}
\DeclareCaptionFormat{cont}{#1 (cont.)#2#3\par}
\usepackage[pagebackref,breaklinks,colorlinks]{hyperref}

\usepackage[capitalize]{cleveref}
\crefname{section}{Sec.}{Secs.}
\Crefname{section}{Section}{Sections}
\Crefname{table}{Table}{Tables}
\crefname{table}{Tab.}{Tabs.}

\def\cvprPaperID{9930} % *** Enter the CVPR Paper ID here
\def\confName{CVPR}
\def\confYear{2022}

\begin{document}

%%%%%%%%% TITLE - PLEASE UPDATE
\title{Video2StyleGAN: Encoding Video in Latent Space for Manipulation}

\author{Jiyang Yu\textsuperscript{1},\ \ \ \ Jingen Liu\textsuperscript{1},\ \ \ \  Jing Huang\textsuperscript{1},\ \ \ \  Wei Zhang\textsuperscript{2},\ \ \ \ Tao Mei\textsuperscript{2}\\
\textsuperscript{1}JD Explore Academy, Mountain View, USA,\\
\textsuperscript{2}JD Explore Academy, Beijing, China\\
{\tt\small \{jiyang173, jingenliu, jinghuang.zhu, wzhang.cu\}@gmail.com, \{tmei\}@jd.com}
% For a paper whose authors are all at the same institution,
% omit the following lines up until the closing ``}''.
% Additional authors and addresses can be added with ``\and'',
% just like the second author.
% To save space, use either the email address or home page, not both
%\and
%Second Author\\
%Institution2\\
%First line of institution2 address\\
%{\tt\small secondauthor@i2.org}
}

\maketitle

%%%%%%%%% ABSTRACT
\begin{abstract}
Many recent works have been proposed for face image editing by leveraging the latent space of pretrained GANs. However, few attempts have been made to directly apply them to videos, because 1) they do not guarantee temporal consistency, 2) their application is limited by their processing speed on videos, and 3) they cannot accurately encode details of face motion and expression.
To this end, we propose a novel network to encode face videos into the latent space of StyleGAN for semantic face video manipulation.
Based on the vision transformer, our network reuses the high-resolution portion of the latent vector to enforce temporal consistency.
To capture subtle face motions and expressions, we design novel losses that involve sparse facial landmarks and dense 3D face mesh. We have thoroughly evaluated our approach and successfully demonstrated its application to various face video manipulations. Particularly, we propose a novel network for pose/expression control in a 3D coordinate system. 
Both qualitative and quantitative results have shown that our approach can significantly outperform existing single image methods, while achieving real-time (66 fps) speed. 
\end{abstract}

%%%%%%%%% BODY TEXT
%\begin{figure}[t]
%\includegraphics[width=0.48\textwidth]{Fig/teaser2.png}
%\caption{Due to the large variation of expressions in videos, existing methods like pSp~\cite{psp} fail to reconstruct correct identity (left and right example) and expressions like closed eyes (middle example). By introducing high-level facial landmarks and mesh losses, our approach is significantly more accurate in reconstruction of identity, pose and expression. Best viewed zoomed in. Please watch our supplementary video \#1 for details.}
%\label{fig:teaser}
%\vspace{-15pt}
%\end{figure}

\section{Introduction}\label{sec:intro}
Generative Adverserial Networks (GANs) has been actively studied and adapted in many computer vision tasks like image/video generation~\cite{gan,cgan,mocogan}, editing\cite{idgi,stylerig,styleflow} and restoration\cite{tecogan,srgan}.
Recent works, including BigGAN~\cite{brock2018large}, ProGAN~\cite{karras2017progressive} and StyleGAN~\cite{stylegan1,stylegan2,karras2020training,karras2021alias}, focus on high-fidelity image synthesis.
Among these works, StyleGAN is widely used in image editing, since its disentangled latent space $\mathcal{W}$ is suitable for manipulating image attributes.
Its latent space is often extended to $\mathcal{W}+$ by varying the input to the hierarchical structure of StyleGAN (ranging from lower to higher resolution layers), providing more flexibility in controlling the output image.
To edit an existing image in $\mathcal{W}+$, its equivalent latent vector must be found so that the pretrained StyleGAN can reconstruct the input image.
This process is called GAN inversion in the literature.
GAN inversion and latent editing for single image have been explored in many works, including optimization~\cite{image2stylegan, image2stylegan++} and deep-learning~\cite{psp,e4e,restyle,simple}.

\begin{figure}[t]
    \centering
    \includegraphics[width=0.48\textwidth]{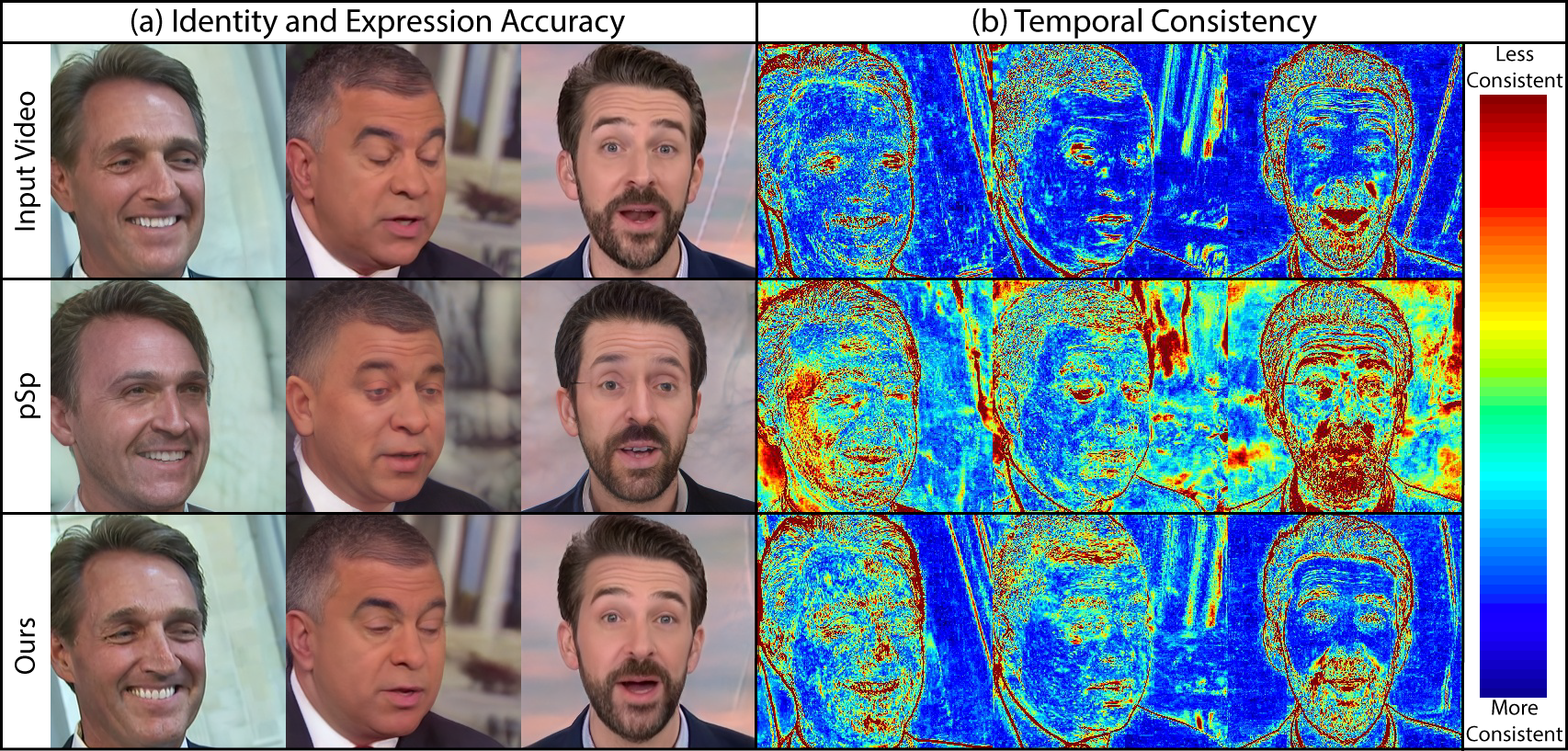}
    \caption{\textit{(a) Due to the large variation of expressions in videos, existing methods like pSp~\cite{psp} fail to reconstruct correct identity (left and right example) and expressions like closed eyes (middle example). (b) Temporal consistency is also not guaranteed in single image GAN inversion methods. By introducing high-level facial landmarks and mesh losses, our approach performs significantly better in reconstruction of identity, pose and expression while maintaining temporal consistency. Best viewed zoomed in. Please watch our supplementary video \#1 for details.}}
    \label{fig:teaser}
\end{figure}

On the other hand, although many sophisticated video editing algorithms are developed~\cite{obama, face2face,wang2021one,zhou2021pose} due to the demand in many real-world applications, few have discussed the possibility of video editing in latent space.
One important advantage of latent space video editing is that, leveraging the prior knowledge from pretrained StyleGAN, complex facial structures do not need to be specifically considered, e.g. generating teeth for the edited lips. In this paper, we aim at solving the video-based GAN inversion problem for face video manipulation in the StyleGAN latent space. To solve this problem, 
%begin{comment}
%\jing{not follow the advantage. do you mean pretrained styleGAN will not generate teeth on lips?}
%\end{comment}
there are three challenges that need to be handled.

First, video encoding and editing require temporal consistency.
Any previous image-based GAN inversion method inevitably suffers from inaccuracy in the encoding.
Without considering temporal information, these inaccuracies can lead to visual jitters in facial appearance, lighting and background in videos. 
Figure~\ref{fig:teaser}(b) shows the accumulated RMSE magnitude map among 6 spatially aligned neighbor frames in the video.
It can be observed that image GAN inversion methods like pSp~\cite{psp} yield temporally inconsistent results (2\textsuperscript{nd} row).
%The second row in Figure~\ref{fig:teaser}(a) shows an example of single frame encoding without considering temporal consistency.
%Note that the appearance change of the face over time.
The temporal inconsistencies are usually difficult to model, % in the image domain.
because the difference caused by jitters is often much smaller than that caused by the video content changes.
Inspired by the observation that the majority of the visual jitters reside in the higher resolution layers of StyleGAN, 
we design a novel video encoding network that reuses the higher resolution portion of the latent vector from the previous frames to reduce visual jitters.
Specifically, our encoding network adapts the vision transformer (ViT)~\cite{vit} structure, which enables easy incorporation of previous frame information. Basically, the network takes the current frame and the latent vector of the previous frame as inputs. It then infers the low resolution layers of latent vector,  
%Given the latent vector of the previous frame, we use a projection network to embed it as an additional token similar to the patch tokens in ViT.
%The video encoding network is only responsible for inferring the low resolution layers of latent vector, 
while the high resolution layers remain the same as the previous frame.
Our approach successfully avoids visual jitters and achieves good visual temporal consistency in the resulting video (see 3\textsuperscript{rd} row in Fig.~\ref{fig:teaser}(b) and supplementary video \#2 ).
%We recommend readers to watch our supplementary video \#2 for best view.
%We will discuss the details of our design in Sec.~\ref{subsec:framework}.

Second, the speed of video-based GAN inversion should be fast enough for real applications.
Optimization based methods~\cite{image2stylegan, image2stylegan++,idgi} usually require hundreds of iterations to encode a single image, which is prohibitive to live stream video and long videos.
Although learning based methods~\cite{psp,e4e,restyle} are faster thanks to the adaptation of hierarchical deep convolutional structures, they are still far from real-time. In contrast, our proposed transformer-based framework has an average 15ms/frame (66 fps, see Table.~\ref{tab:quant}) speed in the GAN inversion, which is the first method that achieves real-time performance to our best knowledge.
%In Table.~\ref{tab:quant}, we will compare the inference time of existing GAN inversion methods.

Last, but most important, subtle expressions and motions like eye blinking and dynamic wrinkles, are critical for reconstructing a high-fidelity face video.
However, the accuracy of poses and expressions has not been discussed and evaluated in existing GAN inversion works.
These works usually have a strict requirement that the input must reside in the same domain of StyleGAN generated images, with faces aligned to the center.
One may argue that face alignment can be done as a pre-processing. Nevertheless, various facial expressions that are not the training set of StyleGAN are still challenging for these works.
%Moreover, existing single image GAN inversion has strict requirement that the face in the input image must be aligned to the center.
%As a result, their methods are limited by the accuracy of facial landmark detection and usually fail catastrophically if any part of the face is not aligned properly.
As a result, these methods usually fail catastrophically in videos because video faces are not aligned properly. As examples shown in Figure~\ref{fig:teaser}(a), the image method pSp~\cite{psp} is less accurate in identity (left and right example) and expression (closed eyes in the middle example).
%In terms of expression accuracy, it could not handle closed eyes (middle example).% which is common in videos.
It also produces ambiguous results, like partial glasses in the right example.
To solve the alignment problem, in our work, we introduce a novel sparse face landmark loss and a novel dense 3D face mesh loss (Sec.~\ref{subsec:loss}).
Note that although explicit 2D facial landmarks are often used in face manipulation works~\cite{nitzan2020face,zhu2021one}, our approach differs from other methods by leveraging raw feature maps of a landmark detector to implicitly guide the video encoding network with high-level face pose and expression information, yielding significantly more accurate GAN inversion than existing methods (See Fig.~\ref{fig:teaser} and Sec.~\ref{subsec:invert}).

To summarize, our contribution includes:

\textbf{\textbullet} We design a novel video-based GAN inversion network that leverages the continuity of video frames (Sec.~\ref{subsec:framework}). To the best of our knowledge, our work is the first GAN inversion model that considers temporal consistency and is specifically designed for videos.

\textbf{\textbullet} Our efficient network design makes inversion fast enough for video applications. Our approach achieves real-time performance (66 FPS) and is significantly faster than existing GAN inversion methods (Table.~\ref{tab:quant}).

\textbf{\textbullet} We introduce a novel sparse face landmark loss and a dense 3D face mesh loss, which enables accurate inversion of faces with arbitrary pose and expression. %(Sec.~\ref{subsec:loss}).

\textbf{\textbullet} We successfully exhibit various video manipulations and propose a novel network for pose/expression control in 3D coordinates for face videos.

%\begin{itemize}
%  \item We design a novel video GAN inversion network that leverages the continuity of video frames (Sec.~\ref{subsec:framework}). To the best of our knowledge, our work is the first GAN inversion model that considers temporal consistency and is specifically designed for videos.
%  \item Our efficient network design makes inversion fast enough for video applications. Our approach achieves real-time performance (66fps) and is significantly faster than existing GAN inversion methods (Table.~\ref{tab:quant}).
%  \item We introduce novel sparse face landmark loss and dense 3D face mesh loss, which enables accurate inversion of faces with arbitrary pose and expression (Sec.~\ref{subsec:loss}).
%  \item We successfully exhibit various video manipulations and particularly propose a novel pose/expression control network for face videos. 
  %\item We propose a novel face video dataset that can be used in a variety of tasks, including StyleGAN inversion and pose editing of face videos.
%\end{itemize}

\section{Related Work}\label{sec:related}
\noindent\textbf{GAN Inversion} 
is a relatively new concept in vision tasks.
Early ideas use it to quantify model collapse in GAN~\cite{gansee} or edit general image to obey high-level semantics~\cite{ganpaint}.
StyleGAN and its subsequent works~\cite{stylegan1,stylegan2,karras2020training,karras2021alias} enables high fidelity image generation.
Motivated by StyleGAN, many works utilize its disentangled latent space for image manipulation.
There are two typical approaches: optimization based and learning based.
Image2StyleGAN~\cite{image2stylegan,image2stylegan++} optimize the input latent vector so that the input image can be reconstructed by a pretrained StyleGAN.
Learning based approach pSp~\cite{psp} and its subsequent work e4e~\cite{e4e} train a pyramid network structure to infer the latent vectors at different spatial resolutions.
ReStyle~\cite{restyle} introduces an iterative refinement mechanism to pSp and e4e, leading to improved results.
Beyond image domain accuracy, many works also argue that GAN inversion results must remain editable in the latent space.
Following this line, works like IDInvert~\cite{idgi} and Editing in Style~\cite{editinstyle} focus on discovering semantics in the StyleGAN latent space.

\begin{figure*}[htb]
    \centering
    \includegraphics[width=\textwidth]{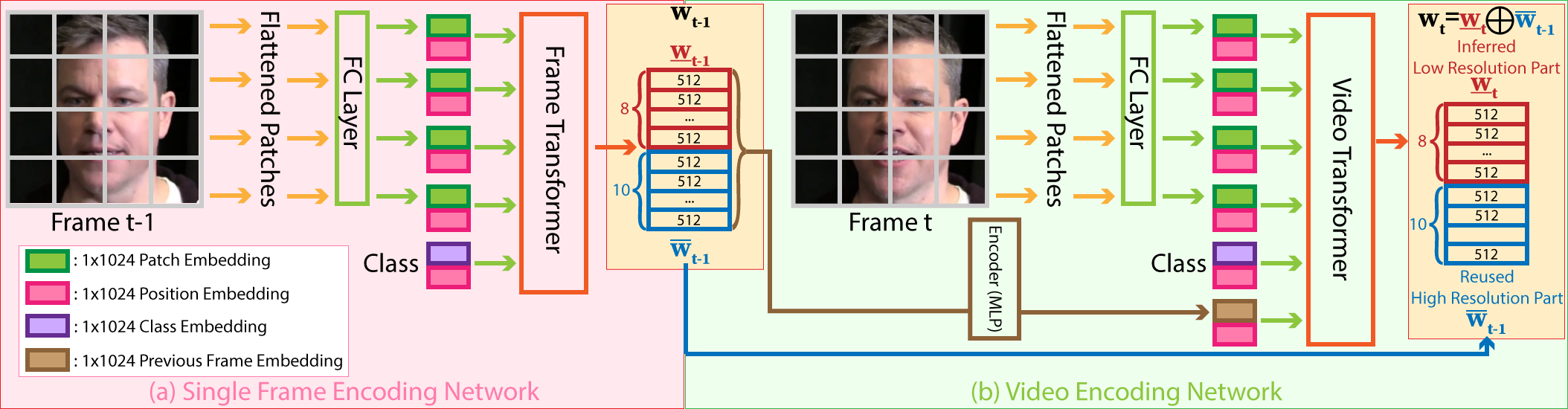}
    \caption{\textit{The structure of our video encoding network. (a) Adapting the recent vision transformer~\cite{vit}, we first regress the latent vector $\mathbf{w}_{t-1}$ for the first frame. (b) In the video encoding network, we encode $\mathbf{w}_{t-1}$ as an extra embedding for the attention, and only infer $\underline{\mathbf{w}}_t$, the low resolution part of the latent vector. We reuse the high resolution part from the previous frame $\overline{\mathbf{w}}_{t-1}$ to avoid appearance jitters.}}
    \label{fig:network}
\end{figure*}

\noindent\textbf{Semantic Manipulation in Latent Space.} The success of GANs on producing photo-realistic images has brought increasing attention to understanding the latent space of pretrained GANs, which is usually considered as Riemannian manifold~\cite{kuhnel2018latent,arvanitidis2017latent}. Some early works~\cite{perarnau2016invertible,upchurch2017deep,radford2015unsupervised} have tried simple arithmetic operations or interpolation of synthesis images in latent space to generate new images. To semantically control the image generation, one solution is to identify the interpretable direction of image transformation and then navigate along this direction in the latent space for a given attribute, such as age, expression, and lighting~\cite{shen2020interpreting,abdal2021styleflow,denton2019detecting,goetschalckx2019ganalyze,jahanian2019steerability}. As a result, the control is to characterize more or less of an image attribute. Another interesting work by
%For example, Shen et al. ~\cite{shen2020interpreting} studies how different semantics are encoded in latent space and how to disentangle multiple semantics of face images. 
Jahanian et al.~\cite{jahanian2019steerability} is to explore the ``steerability’’ of GANs in latent space to conduct simple transformations such as camera movements and image color changes. On the other hand, some recent works~\cite{voynov2020unsupervised,harkonen2020ganspace,shen2021closed} try to identify the manipulation directions in an unsupervised way by leveraging eigenvector decomposition. But, they need to manually select meaningful directions.

One application of latent space manipulation is semantic face editing, which aims at manipulating a target facial attribute while keeping other information unchanged. To this end, some works~\cite{chen2016infogan,tran2017disentangled,bao2018towards,lample2017fader,xiao2018elegant,yin2017towards,donahue2017semantically} propose various strategies to train new disentangled GAN models for semantic editing. Recently, more attentions have been shifted to conduct face editing inside the latent space of learned GAN models~\cite{psp,e4e,idgi,editinstyle,shen2020interpreting,image2stylegan}. Different from face image editing, we aim at learning a disentangled space for face video manipulation, which was rarely explored. As discussed in Sec.~\ref{sec:intro}, our proposal can solve all three challenges in face video manipulation.

\section{Embedding Video in Latent Space}\label{sec:encode}
\subsection{Background}
The StyleGAN~\cite{stylegan1,stylegan2,karras2020training,karras2021alias} %is a GAN framework that generates high resolution image from random noise input.
generator consists of a 9-level pyramid structure operating on spatial resolution from $4\times4$ to $1024\times1024$.
Using adaptive instance normalization (AdaIN)~\cite{adain}, the 18 512-dimension vectors, often called styles, are injected into different resolution levels to control the appearance of the output image.
The lower-resolution layers control the high-level layout/shape of the face, while the higher-resolution layers control the face details. 
The space of these vectors is denoted as $\mathcal{W}+\in \mathbb{R}^{18\times 512}$.
In our work, a pretrained StyleGANv2~\cite{stylegan2} generator is treated as a black-box renderer.
Our goal is to project each frame of the video into a latent vector $\mathbf{w}$ in $\mathcal{W}+$, so that the pretrained StyleGAN reconstructs the exact input frame.

As discussed in Sec.~\ref{sec:intro}, video GAN inversion has three challenges:
\textbf{I.} Keeping the temporal consistency between consecutive output frames; % should be guaranteed;
\textbf{II.} Running at real-time; % should be fast enough for practical video applications;
\textbf{III.} Retaining subtle motion and expression. % change needs to be reconstructed properly.
We will try to solve these challenges in this work.
%In the following sections, we will introduce our solution to these challenges.

%In Sec.~\ref{subsec:framework}, aiming to tackle challenges 1) and 2), we introduce the structure of our StyleGAN inversion network for face video. In Sec.~\ref{subsec:loss}, we discuss the design of loss functions aiming to tackle challenge 3). We also collect a face video dataset in order to train the StyleGAN inversion network for face video. In Sec.~\ref{subsec:training}, we describe the details of our dataset and the training procedure of our network.

\subsection{Our Framework}\label{subsec:framework}
Inspired by the structure of StyleGAN, many single image GAN inversion works~\cite{psp, e4e, simple} use a hierarchical structure, which encodes the input image into the latent vectors. 
For the video GAN inversion, however, in order to follow the same structure, we need to adapt some recurrent structures like convolutional LSTM to incorporate temporal information. 
But, such designs not only make the network difficult to train due to the high GPU memory demand, but also is inefficient. % for video GAN inversion task. 
Unlike existing GAN inversion works, we adapt a vision transformer (ViT)~\cite{vit} in our framework.
It unwraps the input image into patches and embeds them into a low dimensional space. Its benefit is twofold: 1) It embeds neighbor frames into the same space, which enables involving temporal information at minimum cost; 2) It is efficient in speed. So, it can solve Challenge I and II simultaneously. 
%The benefit of this structure is that neighbor frame information can be embedded into the same space and assist the inversion of the current frame.
%This makes the network able to involve temporal information at minimum additional cost, solving the challenge 1) and 2) simultaneously.

For simplicity, we use ``encoding'' to represent GAN inversion in the rest of our discussion. 
Fig.~\ref{fig:network} depicts the overview structure of our framework, which consists of the Single Frame Encoding Network (FrEN) and Video Encoding Network (ViEN). The FrEN is trained to encode each frame separately. It directly regresses the $\mathcal{W}+$ vector $w\in \mathbb{R}^{18\times512}$ from the class token of ViT as shown in Fig.~\ref{fig:network}(a). As aforementioned, encoding a video frame-by-frame will cause visual jitters in videos. To remove jitters, the ViEN also takes the latent vector of the previous frame as input, other than the current frame. The first frame of a video is encoded by the FrEN. 

%Note that since training video encoding network requires a known latent vector of the previous frame, a single frame encoding network is needed to infer the latent vector of the initial frame.
%For videos, encoding in a frame-by-frame fashion cannot guarantee temporal consistency.

%\begin{figure}[t]
%\includegraphics[width=0.48\textwidth]{Fig/consistency2.png}
%\caption{Visualization of temporal consistency in the result video. We align 6 neighbor frames to the center frame (1\textsuperscript{st} column). We accumulate the RMSE with the center frame, and plot the error maps for three cases at the same scale for comparison. The average RMSE value is shown at the bottom of each error map (color blue indicates smaller error). Please watch our supplementary video \#2 for direct observation.}
%\label{fig:consistency}
%\end{figure}

\begin{figure}[t]
    \centering
    \includegraphics[width=0.48\textwidth]{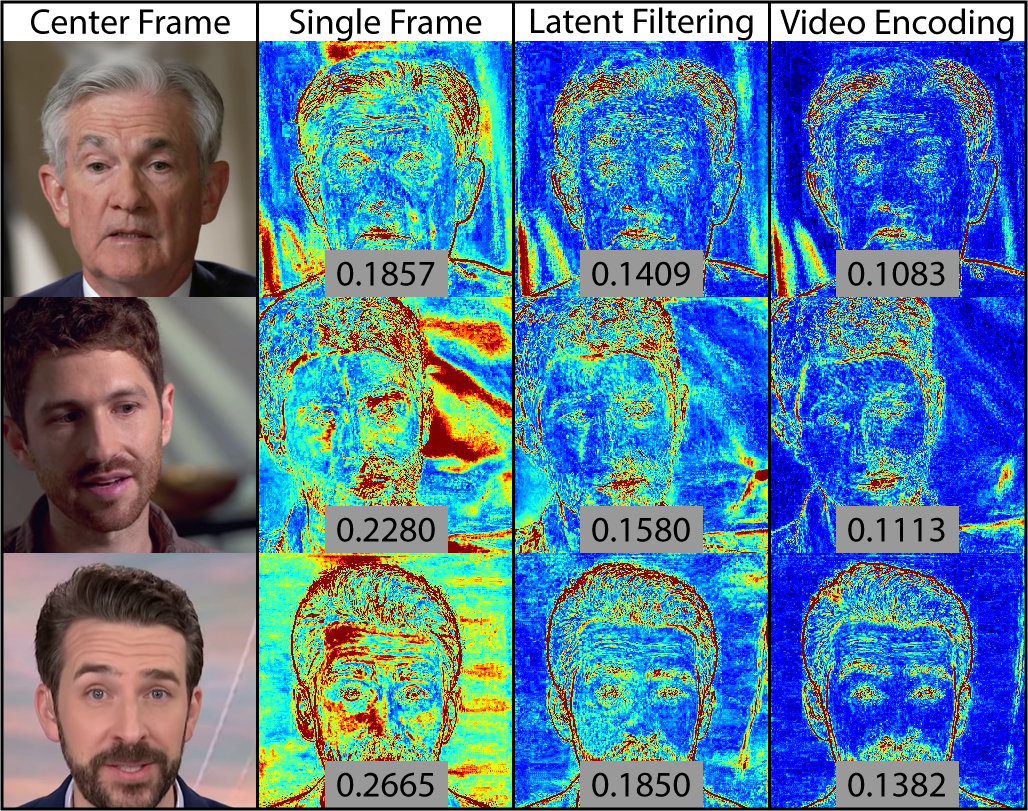}
    \caption{\textit{Visualization of temporal consistency in the result video. We align 6 neighbor frames to the center frame (1\textsuperscript{st} column). We accumulate the RMSE with the center frame, and plot the error maps for three cases at the same scale for comparison. The average RMSE value is shown at the bottom of each error map (color blue indicates smaller error). Please watch our supplementary video \#2 for direct observation.}}\label{fig:consistency}
\end{figure}

To illustrate the importance of temporal consistency, we visualize the jitters in Fig.~\ref{fig:consistency}. We align 6 reconstructed neighbor frames to their center frame and compute the accumulated root mean square error (RMSE) between the aligned neighbor frame and the center one. 
The total RMSE is shown at the bottom of each error map. A larger value (red color) indicates the reconstructed video has more visual jitters. 
In the 2\textsuperscript{nd} column, although the individual frames are reasonably reconstructed from the latent vectors, single frame encoding results in large visual jitters (e.g. lighting and facial appearance).
Such jitters are obvious in human eyes, please watch our supplementary video \#2 for better effects.
In our experiment, we observe that these appearance jitters originate from the higher resolution layers in StyleGAN.
So we try to use a Gaussian filter to smooth the 9-18 layers of $\mathbf{w}$ along the time dimension. The results in the 3\textsuperscript{rd} column of Fig.~\ref{fig:consistency} indicate the Gaussian smoothing can alleviate the jitters to some extend.
However, the naive smoothing in the latent space can also average out the motion of the face as shown in the supplementary video \#2. This may due to
that the face pose is not completely disentangled in the $\mathcal{W}+$ space.
To remove jitters, we need to fix the high resolution layers in $\mathbf{w}$, and also appropriately compensate the face motion caused by directly using high resolution layers from the previous frame.

Let us denote the high and low resolution layers of the latent vector $\mathbf{w}$ as $\overline{\mathbf{w}}\in \mathbb{R}^{10\times512}$ and $\underline{\mathbf{w}}\in\mathbb{R}^{8\times512}$, respectively.
As shown in Fig.~\ref{fig:network}(b), the input of the ViEN is the current frame and the latent vector $\mathbf{w}_{t-1} \in \mathbb{R}^{18\times512}$ from the previous frame.
Using an MLP projector, we encode the latent vector from the previous frame as an extra embedding input for the ``video transformer''.
The output of the ViEN is the lower resolution layers $\underline{\mathbf{w}}_{t}\in \mathbb{R}^{8\times512}$.
We obtain the encoded latent vector as $\mathbf{w}_t=\underline{\mathbf{w}}_t \oplus \overline{\mathbf{w}}_{t-1}$, where $\oplus$ represents concatenation.
Please note that at inference time, $\mathbf{w}_{t-1}$ is the output of ViEN of the previous frame. The first frame of a video is actually encoded by the FrEN.  
%Concatenating $\underline{\mathbf{w}}_t$ and $\overline{\mathbf{w}}_{t-1}$, we obtain the encoded latent vector $\mathbf{w}_t$ for the current frame.
The final reconstructed frame $\widehat{I}_t$ can be rendered by a pretrained StyleGAN $G$ as: $\widehat{I}_t=G(E(I_t,\mathbf{w}_{t-1}))$,
%The entire process can be written as:
%\begin{equation}
%    \widehat{I}_t=G(E(I_t,\mathbf{w}_{t-1}))
%\end{equation}\
where $E$ denotes the video encoding network.

\begin{figure*}[t]
    \centering
    \includegraphics[width=\textwidth]{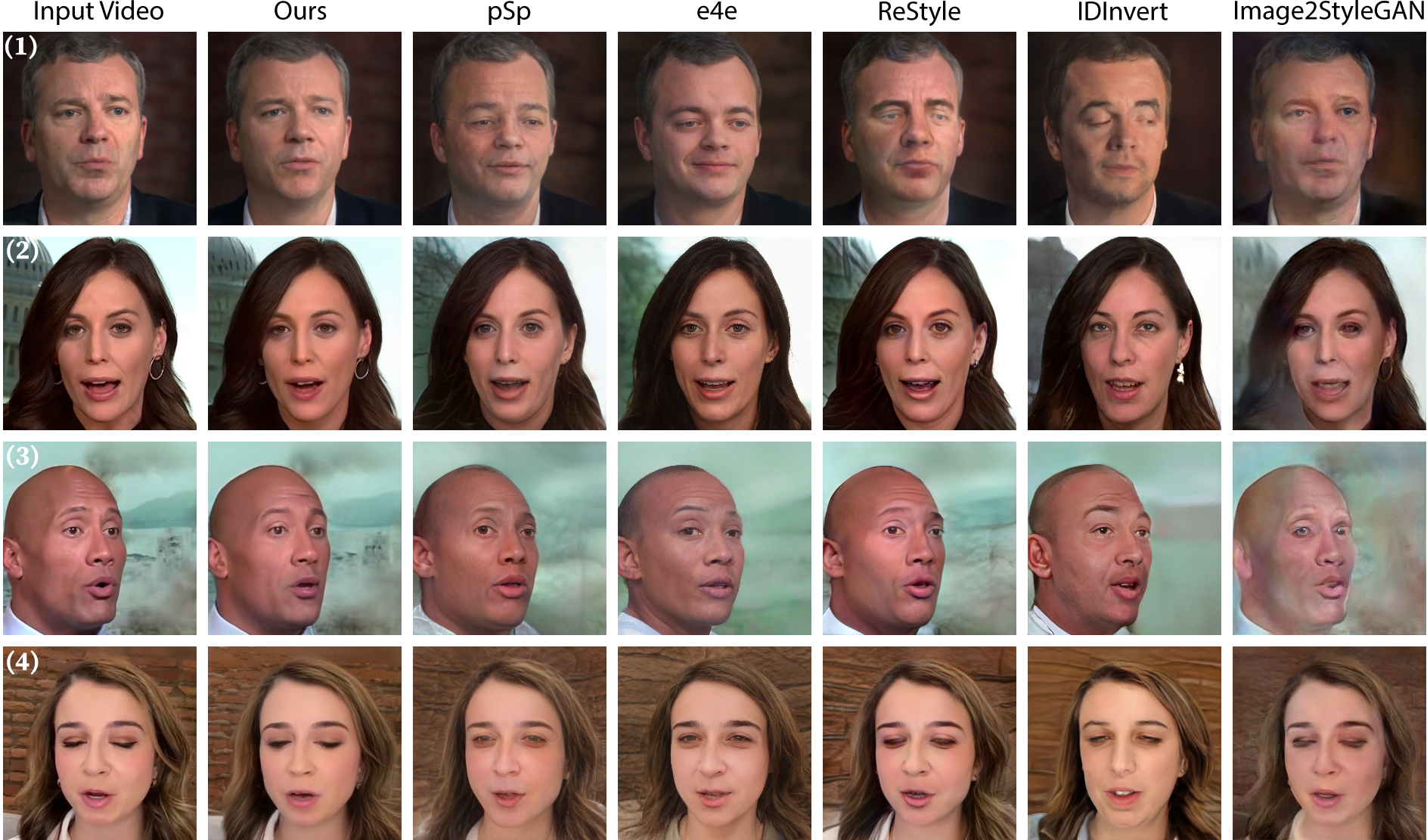}
    \caption{\textit{Qualitative comparison of our method with pSp~\cite{psp}, e4e~\cite{e4e}, ReStyle~\cite{restyle}, IDInvert~\cite{idgi} and Image2StyleGAN~\cite{image2stylegan}. Example (1)-(4) are sorted according to the difficulty of the facial expression reconstruction, from easy to difficult. Example (1) have a neutral expression; (2) and (3) are typical talking expressions; (4) contains eye blinking which is challenging to the comparison methods. Best viewed zoomed in. Please watch our supplementary video \#3 for details.}}
\label{fig:inversion}
\end{figure*}

\subsection{Loss Functions}\label{subsec:loss}
Having guaranteed temporal consistency by reusing the high resolution portion of the latent vector (Sec.~\ref{subsec:framework}), we design losses solely focusing on high fidelity reconstruction of the face for an individual frame.
Image GAN inversion works have proved the effectiveness of $L_2$ and perceptual loss, which directly evaluate the similarity between the input and reconstructed image.
Let the current frame be $I_t$ and the latent vector of last frame as $\mathbf{w}_{t-1}$.
The $L_2$ loss in our training is then defined as:
\begin{equation}\label{eqn:l2}
    L_2(I_t,\mathbf{w}_{t-1})=\left \| I_t-\widehat{I}_t \right \|_2.
\end{equation}
LPIPS~\cite{lpips} loss compares extracted intermediate features via a pretrained backbone network to evaluate the perceptual similarity between two images.
We apply LPIPS loss to further enhance the image quality:
\begin{equation}\label{eqn:lpips}
    %\begin{aligned}
    L_{\{vgg,alex\}}(I_t,\mathbf{w}_{t-1})= P_{\{vgg,alex\}}(I_t,\widehat{I}_t),%\\
    %L_{alex}(I_t,\mathbf{w}_{t-1})&= P_{alex}(I_t,\widehat{I}_t),
    %\end{aligned}
\end{equation}
where both VGG and AlexNet are used as the backbone.% of the LPIPS loss.

Most importantly, as discussed in Challenge III, being able to accurately encode the facial expression and motion is critical for video encoding.
This did not gain attention in previous works, since human vision is not sensitive to the subtle discrepancy in still comparisons.
However, in videos, people tend to easily observe inaccurate face motions (e.g. eye blink).
To tackle this challenge, we seek to leverage high-level face information, i.e. sparse facial landmarks and dense 3D face mesh, to guide the network to implicitly capture the face motion. As a result, we propose a sparse facial landmark loss, which enforces the reconstructed face to have the same detected landmarks as the input face. 
This loss is defined based on the 2D FAN~\cite{fan} landmark detector:
\begin{equation}\label{eqn:landmark}
    L_f(I_t,\mathbf{w}_{t-1})=\left \| \mathbf{F}(I_t)-\mathbf{F}(\widehat{I}_t) \right \|_2.
\end{equation}
FAN estimates the position of 68 facial landmarks using $\argmax$ on the 68 channel heatmaps inferred by the network, which is not differentiable.
Therefore, in Eqn.~\ref{eqn:landmark}, we define the landmark loss over the intermediate heatmaps $\mathbf{F}\in\mathbb{R}^{68\times H \times W}$ ($H$ and $W$ are the frame height and width) to evaluate the landmark similarity between the input frame and the reconstructed one. It is worth noting that, our way to define the loss differs from most previous face landmark losses \cite{nitzan2020face,zhu2021one} which utilize L2 loss
on explicit 2D facial landmarks. The continuous feature maps can preserve the original response from a face landmark detector, which provides more implicit guidance for the training of the video encoding network.

In addition, some face motions (e.g., lifting eyebrows) cannot be accurately described by sparse facial landmarks.
Hence, we further use 3DDFAv2~\cite{3ddfa} to model these face geometry details.
Based on the dense 3DMM~\cite{bfm} mean face mesh, 3DDFAv2~\cite{3ddfa} estimates the shape and expression parameters so that the geometry of the face mesh matches the input image.
Let $\mathbf{M}$ and $\mathbf{V}$ denote the mean and result 3D face vertices. The relation between the input frame and the result 3D mesh can be written as:
\begin{equation}
    \mathbf{V}(I)=\mathbf{R}(I)(\mathbf{M}+\mathbf{A}_{id}\mathbf{\alpha}(I)+\mathbf{B}_{exp}\mathbf{\beta}(I))+\mathbf{T}(I),
\end{equation}
where $\mathbf{A}_{id}$ and $\mathbf{B}_{exp}$ are the principal components of identity and expression provided by 3DMM; $\mathbf{\alpha}$, $\mathbf{\beta}$, $\mathbf{R}$ and $\mathbf{T}$ are the identity, expression and rigid transformation parameters inferred by 3DDFAv2. 
We then define the dense 3D loss as:
\begin{equation}\label{eqn:dense}
    L_v(I_t,\mathbf{w}_{t-1})=\left \| \mathbf{V}(I_t)-\mathbf{V}(\widehat{I}_t) \right \|_2.
\end{equation}
As discussed in Sec.~\ref{subsec:invert}, our sparse facial landmark loss and dense 3D loss enable to handle arbitrary face pose and expression without explicitly face alignment as a pre-processing. 
The final loss function of our network can be written as:
\begin{equation}\label{eqn:total}
    L=L_2+\lambda_{p}(L_{vgg}+L_{alex})+\lambda_{f}L_{f}+\lambda_{v}L_{v},
\end{equation}
where $\lambda_{p}$, $\lambda_{f}$ and $\lambda_{v}$ are regularization values for the corresponding losses.
%The values of these regularization values will be discussed in Sec.~\ref{subsec:training}.
%In Sec.~\ref{subsec:abalation}, we will further demonstrate the effectiveness of the losses defined in this section. 

\subsection{Training}\label{subsec:training}
To train the video encoding network, we collected a face video dataset ( see Sec.~\ref{sec:dataset}).
The training procedure consists of two stages:
First, we set $\lambda_{p}=0.8$, $\lambda_{f}=100$ and $\lambda_{v}=0$ and train the network for 90,000 iterations.
In our experiment, we observed that the result is initially of poor quality, and 3DDFAv2~\cite{3ddfa} mostly returns abnormal face geometries.
Therefore, we only apply the soft constraint provided by 2D FAN~\cite{fan} in this stage.

Second, from the 90,000\textsuperscript{th} iteration, we set $\lambda_{p}=0.8$, $\lambda_{f}=0$ and $\lambda_{v}=0.0001$ and train until converge.
In this stage, the quality of the encoded frames becomes more suitable for 3DDFAv2. 
We apply a relatively small weight on the dense 3D loss, so that the training is robust to failure cases in 3DDFAv2.

Since the training of our video encoding network requires knowing the previous frame latent vector, we train the FrEN first using the above procedure.
After the FrEN converges, we train the ViEN using two consecutive video frames drawn from our face video dataset.
The latent vector of the first frame is inferred by the pretrained single frame encoding network, and the ViEN is trained using the second frame and the latent vector of the first frame.
For all training steps, we use the Ranger optimizer following pSp~\cite{psp}. The learning rate is set to 0.0001.
\begin{figure*}[t]
    \centering
    \includegraphics[width=\textwidth]{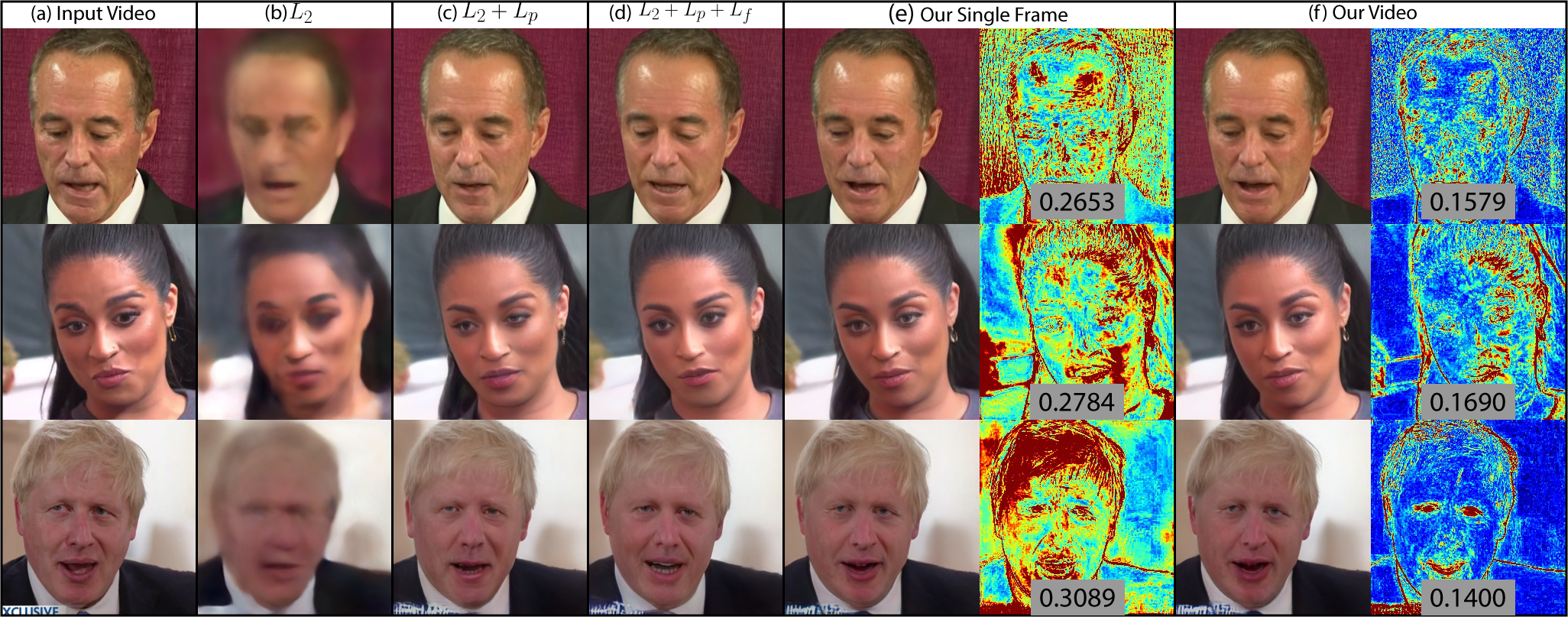}
    \caption{\textit{Effects of training our video encoding network with different loss setups. In column (b), (c) and (d), we gradually adding $L_2$, $L_p$ and $L_f$ respectively. Columns (e) and (f) show our single frame and video encoding results with the complete loss setup (Eqn.~\ref{eqn:total}), and the corresponding accumulated RMSE maps are shown on the right (color blue indicating smaller error). Best viewed zoomed in. Please watch our supplementary video \#4 for details.}}
\label{fig:ablation-loss}
\end{figure*}
\section{Experiments}\label{sec:result}

\subsection{Dataset}\label{sec:dataset}
To train the video GAN inversion network, we collected a dataset containing 16300 face videos from the Internet.
The number of frames in each video is between 50 and 100. 
For each frame, we crop the face region with 256x256 resolution.
The total number of individual frames is 1,648,849.
The sources are TV News and interviews, which cover a wide range of gender, ethnicity, appearance and expression.
These variations play an important role in training the pose-aware video GAN inversion and the pose editing network.

\subsection{Video StyleGAN Inversion}\label{subsec:invert}
For the GAN inversion task, we seek to obtain a sequence of latent vectors so that the pretrained StyleGAN can reconstruct the input video using the latent vectors.
We compare our method with learning-based image GAN inversion methods pSp~\cite{psp}, e4e~\cite{e4e}, Re-Style~\cite{restyle} and optimization based method Image2StyleGAN~\cite{image2stylegan} and IDInvert~\cite{idgi}.
Fig.~\ref{fig:inversion} shows the qualitative comparison.

Example (1) is a relatively neutral expression similar to the images in the StyleGAN training set FFHQ~\cite{stylegan1}.
Example (2) and (3) are typical frames in a talking face video with different expressions compared to FFHQ.
Example (4) contains significant expressions, i.e., eye blinking, which is very challenging for existing GAN inversion methods.
In general, existing GAN inversion methods fail to properly reconstruct the face in the video due to the variety of facial expressions and head poses.
Even if they are able to generate reasonable-looking faces, the result typically contains wrong identity (pSp, e4e and IDInvert in example (1), (3) and (4)) and inaccurate expression (example (4)).
Optimization based approach Image2StyleGAN is able to reconstruct identity and expression to some extent, but it may take a significantly longer time to converge to a better result (See Table.~\ref{tab:quant}), and also could not guarantee the visual quality and temporal consistency.
Since we introduce losses (Eqn.~\ref{eqn:landmark} and Eqn.~\ref{eqn:dense}) to implicitly guide the video encoding network, our model automatically perceives the location of the face and is able to handle arbitrary pose and expression.
We recommend readers to watch our supplementary video \#3 for better comparison, since some attributes of the videos (for example, face motion and temporal consistency) can only be observed in dynamics.

To quantitatively evaluate the performance, we collect a 38 face video test set.
In the upper part of Table~\ref{tab:quant}, we evaluate the accuracy of video GAN inversion using a variety of metrics.
The PSNR and SSIM measure the image domain similarity.
We also include Fréchet inception distance (FID) for the perceptual similarity, since PSNR and SSIM do not evaluate the general fidelity of the reconstructed image.
The facial landmark error (LM) is the mean squared error (MSE) between the facial landmarks of the input video and the result, measuring the accuracy of the face pose and expression.
To evaluate the temporal consistency of the resulting video, we use the accumulated RMSE of 6 aligned neighbor frames defined in Sec.~\ref{subsec:framework}, denoted as TC in Table ~\ref{tab:quant}.

\begin{table*}[t]%[18]{0.65\textwidth}
%\vspace{-20pt}
%\small
\centering
\begin{tabular}{p{0.15\textwidth}||x{0.11\textwidth}|x{0.11\textwidth}|x{0.11\textwidth}|x{0.11\textwidth}|x{0.11\textwidth}|x{0.11\textwidth}}
\hline
             & PSNR$\uparrow$  & SSIM$\uparrow$   & FID$\downarrow$   & LM$\downarrow$ & TC$\downarrow$ & t/frame$\downarrow$    \\ \hline
pSp~\cite{psp}          & 21.64 & 0.6636 & 93.57 & 110.8 & 0.1967  & 84ms       \\ \hline
e4e~\cite{e4e}          & 20.22 & 0.6247 & 100.5 & 118.6 & 0.2066  & 49ms       \\ \hline
ReStyle~\cite{restyle}      & 23.11 & 0.6973 & 91.59 & 84.05&  0.2125  & 354ms      \\ \hline
IDInvert~\cite{idgi}     & 18.92 & 0.5696 & 121.9 & 205.6 &  0.2462 & $\sim$8s   \\ \hline
Im2Style~\cite{image2stylegan}     & 21.13 & 0.7470 & 100.1 & 4.563 & 0.6871  & $\sim$5min \\ \hline
\textbf{Ours}         & \textbf{25.96} & \textbf{0.7872} & \textbf{39.56} & \textbf{2.704} & \textbf{0.1483}  & \textbf{15ms}       \\ \hline \hline
(b) \tiny{$L_2$}           & 25.77 & 0.7655 & 137.2 & 141.4  & 0.1215 &            \\ \hline
(c) \tiny{$L_2+L_p$}         & 23.66 & 0.7363 & 53.14 & 4.325 & 0.1537  &            \\ \hline
(d) \tiny{$L_2+L_p+L_f$}           & 23.25 & 0.7314 & 52.75 & 3.830 & 0.1552  &            \\ \hline
(e) Our Single & 25.52 & 0.7773 & 38.36 & 2.846 &  0.2562 &          \\ \hline
\end{tabular}
\caption{\textit{Quantitative comparison based on PSNR, SSIM, Fréchet inception distance (FID), facial landmark MSE (LM) and temporal consistency (TC). Uparrow indicates higher value is better, downarrow indicates lower value is better. Our method achieves significantly superior performance in videos with a real-time runtime performance (66fps). In the lower part of the table, we also list results of different loss setups and our single frame encoding with complete loss (Eqn.~\ref{eqn:total}).}}\label{tab:quant}

\end{table*} 

In general, existing learning-based single image GAN inversion methods pSp, e4e, ReStyle and IDInvert fail to reconstruct a satisfactory appearance (low PSNR and SSIM, high FID) and expression (high LM) of face videos.
Optimization based method Image2StyleGAN~\cite{image2stylegan} is able to reconstruct facial expression details (high SSIM and low LM), but the overall appearance performance (PSNR and FID) remains similar to the learning-based methods.
Note that the temporal consistency (TC) of the comparison single image methods are all large, especially Image2StyleGAN~\cite{image2stylegan}. 
This indicates that existing methods cannot be generalized to videos directly.
Our method is significantly better in all the metrics, since our GAN inversion network structure guarantees temporal consistency (Sec.~\ref{subsec:framework}) and is trained with face-aware losses like sparse face landmark loss (Eqn.~\ref{eqn:landmark}) and dense 3D face mesh loss (Eqn.~\ref{eqn:dense}).
Also note the per-frame runtime on the right column, our method is the only real-time method that is realistic for video processing, compared to the prohibitive processing time of IDInvert and Image2StyleGAN.
%Note that the speeds of optimization based methods like IDInvert~\cite{idgi} and Image2StyleGAN~\cite{image2stylegan} are prohibitive for real video applications.
%Thanks to our simple network structure (Sec.~\ref{subsec:framework}), our method is the only real-time method. %that achieves real-time performance.

\subsection{Effectiveness of Losses}\label{subsec:abalation}

To demonstrate the effectiveness of the loss design discussed in Sec.~\ref{sec:encode}, we show the comparison of our models trained with different loss setups in Fig.~\ref{fig:ablation-loss}.
In column (b), the results produced by the network trained only with $L_2$ loss (Eqn.~\ref{eqn:l2}) are blurry.
Adding the perceptual loss (Eqn.~\ref{eqn:lpips}) promotes the overall quality of the image in column (c).

However, due to the lack of high-level understanding of face semantics, the accuracy of the face pose(3\textsuperscript{rd} row) and expression (1\textsuperscript{st} and 2\textsuperscript{nd} rows) are not satisfactory.
By applying the sparse facial landmark loss (Eqn.~\ref{eqn:landmark}, our network is able to perceive the facial expression in the video, as shown in column (d).
The dense 3D loss (Eqn.~\ref{eqn:dense}) further guides our network to capture subtle expression, making the final GAN inversion result accurately reconstructs the input (columns (e) and (f)).
Similar to Fig.~\ref{fig:consistency}, we include the accumulated RMSE for the result of single frame encoding and video encoding networks.
By reusing the high resolution part of the latent vector, our video encoding network is able to reduce the appearance jitters compared to single frame encoding.
In the bottom part of Table~\ref{tab:quant}, we also show the quantitative comparisons of these loss setups.
Note that although setup (b) and (c) is temporally consistent (low TC) and achieve better PSNR and SSIM, they produce undesired blurry and inaccurate facial expression (high FID and LM) since they are trained solely on image domain losses.

%\begin{figure*}[t]
%\includegraphics[width=\textwidth]{Fig/pose_network2.png}
%\caption{The structure of our pose/expression editing network.}
%\label{fig:pose-net}
%\end{figure*}

\begin{figure}[t]
    \centering
    \includegraphics[width=0.48\textwidth]{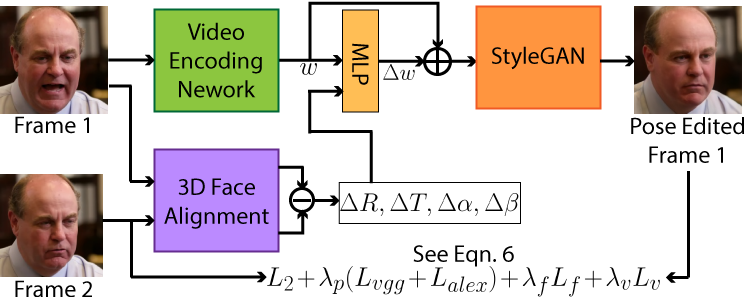}
    \caption{\textit{The structure of our pose/expression editing network.}}
    \label{fig:pose-net}
\end{figure}

\subsection{Pose/Expression Editing}
Due to the complexity of the human face, physically modeling and editing the face appearance \cite{obama, face2face,wang2021one,zhou2021pose} is difficult in general.
One benefit of encoding a video into the latent space is to control the face pose and expression editing, which is important for video generation.
Face pose editing is a challenging task, especially for videos.
Due to occlusion, part of the face is often invisible in certain frames, e.g. ears and teeth.
Adapting the edited face with other regions in the video, e.g. neck and hair, is also difficult.
Traditional methods build sophisticated algorithms for face editing, i.e. learning a warp field to transform high dimensional features~\cite{wang2021one}.
However, these methods could be subject to robustness issues due to the complexity of real-world cases.

\begin{figure*}[t]
    \centering
    \includegraphics[width=\textwidth]{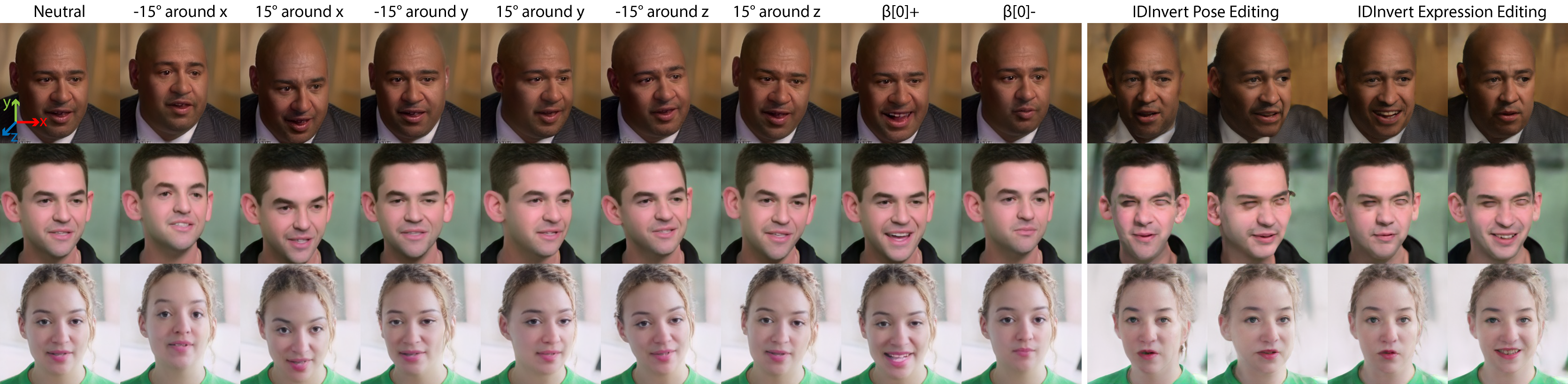}
    \caption{\textit{Pose editing results for videos. Thanks to our face video dataset, we train a pose editing network that enables pose editing in a 3D coordinate system shown on the bottom left of the first row. We can also edit the expression using the parameters defined by the 3DMM face model~\cite{bfm}. Note the high fidelity and flexibility of our pose editing results, compared to existing single image pose and expression editing IDInvert~\cite{idgi} (right part). Best viewed zoomed in. Please watch our supplementary video \#5 for details.}}
\label{fig:pose}
\end{figure*}

On the other hand, since StyleGAN latent space encodes the general face semantics, the rendered face is guaranteed to lie in this space, making the pose/expression editing much easier.
We are able to focus on changing the face pose/expression, but do not need to put extra effort into maintaining the fidelity of the rendering.
Moreover, since temporal consistency is guaranteed by our video encoding network and the same edits are performed on the latent codes, the editing does not need to consider temporal consistency.
Our approach of the network is shown in Fig.~\ref{fig:pose-net}.
Thanks to our face video dataset, we are able to train the pose editing network using the complementary face pose variations in the dataset.
For each iteration, we select 2 frames ($I_1$ and $I_2$) from the same video clip.
To incorporate enough pose and expression difference, $I_1$ and $I_2$ are 10 frames apart.
We use 3DDFAv2~\cite{3ddfa} to estimate shape $\mathbf{\alpha}_1, \mathbf{\alpha}_2$, expression $\mathbf{\beta}_1, \mathbf{\beta}_2$, rotation $\mathbf{R}_1, \mathbf{R}_2$ and translation $\mathbf{T}_1, \mathbf{T}_2$ on $I_1$ and $I_2$.
We concatenate the shape difference $\Delta\mathbf{\alpha}=\mathbf{\alpha}_2-\mathbf{\alpha}_1$, expression difference $\Delta\mathbf{\beta}=\mathbf{\beta}_2-\mathbf{\beta}_1$, rotation difference $\Delta\mathbf{R}=\mathbf{R}_2-\mathbf{R}_1$ and translation difference $\Delta\mathbf{T}=\mathbf{T}_2-\mathbf{T}_1$ as the input of a simple 3-layer MLP, which is responsible for inferring the latent vector displacement $\Delta \mathbf{w}$.
Using our single frame encoding network, we obtain the latent vector for the first frame $\mathbf{w}_1$.
Finally, we use the same training losses (Eqn.~\ref{eqn:total}) to enforce the frame reconstructed from $\mathbf{w}_1+\Delta \mathbf{w}$ to be similar as $I_2$.

\begin{figure*}[t]
    \centering
    \includegraphics[width=\textwidth]{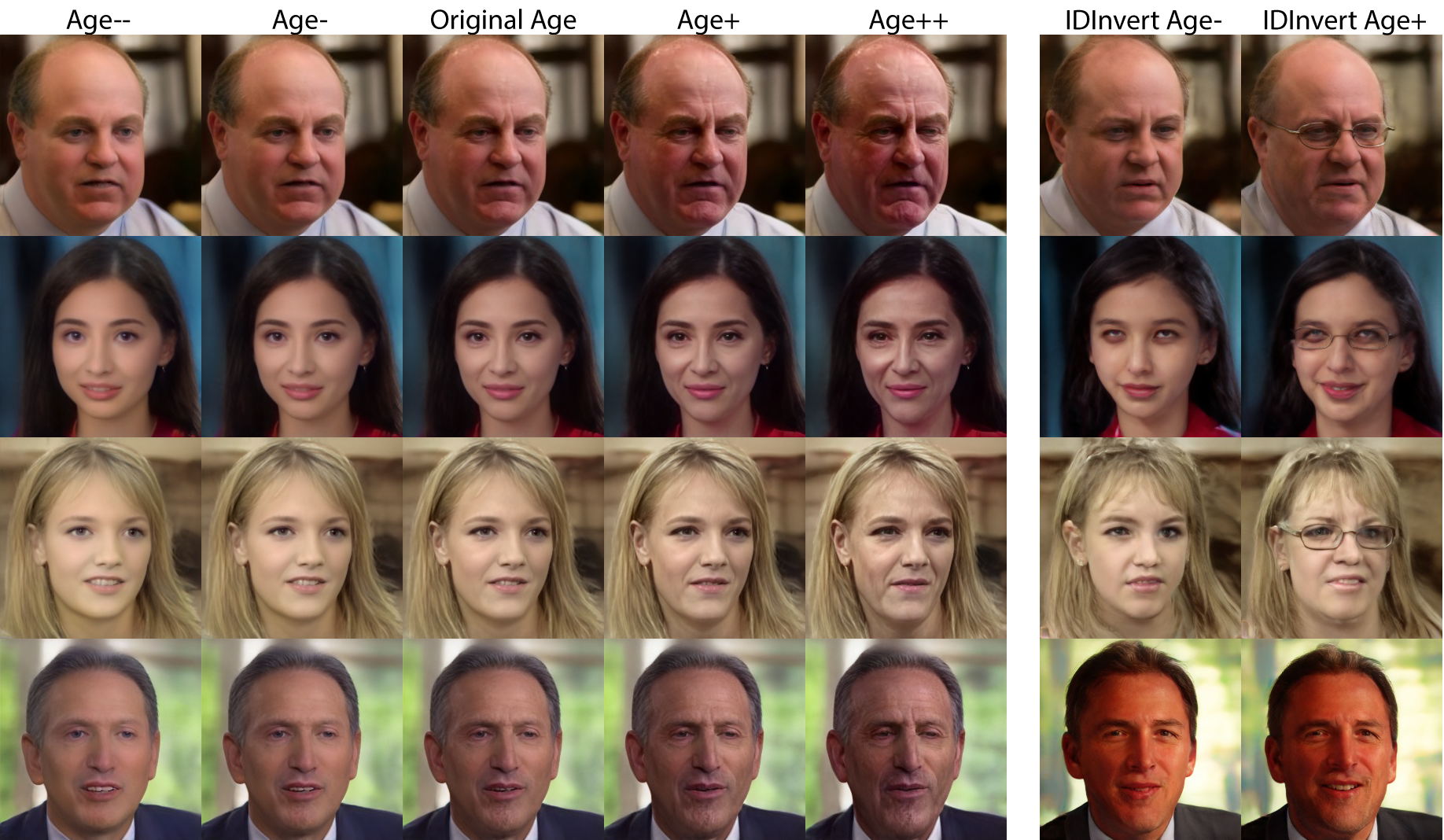}
    \caption{\textit{Age editing results for videos. Left part: the center column shows the original age; the left columns gradually decrease the age; The right columns increase the age. Right part: age editing result of IDInvert~\cite{idgi}. Best viewed zoomed in. Please watch our supplementary video \#6 for details.}}
\label{fig:age}
\end{figure*}

Fig.~\ref{fig:pose} demonstrates examples on pose and expression editing.
Since we define the input based on the 3DMM~\cite{bfm} face model parameters, we are able to edit both pose and expression in a flexible way.
Following the coordinate system shown in the upper left image, the face can be rotated in the 3D coordinates.
The other parts of the frame, e.g. connection with neck, hair and clothes, are automatically handled.
The occlusion relation between face and ears is also handled without any artifacts.
In the expression editing (last two columns), although the teeth are not visible in the input frame, our method is able to generate them without any human intervention.
\textbf{\textit{Since we have shown that existing single image GAN inversion methods cannot handle videos (Sec.~\ref{subsec:invert}), they could not produce satisfactory video editing results.}}
Therefore, due to limited space, we only compare with IDInvert~\cite{idgi} and pSp~\cite{psp} if a pre-trained single image editing model is provided for a task discussed in our paper.
%We also include comparison with single image editing result from IDInvert~\cite{idgi}.
Note that they only support binary editing with a specific attribute, and generate artifacts (first two examples) and change identity (third example).
Please watch our supplementary video \#5 for a better view of the temporal consistency of our results.

\subsection{Appearance Editing}

\begin{figure}[t]
    \centering
    \includegraphics[width=0.48\textwidth]{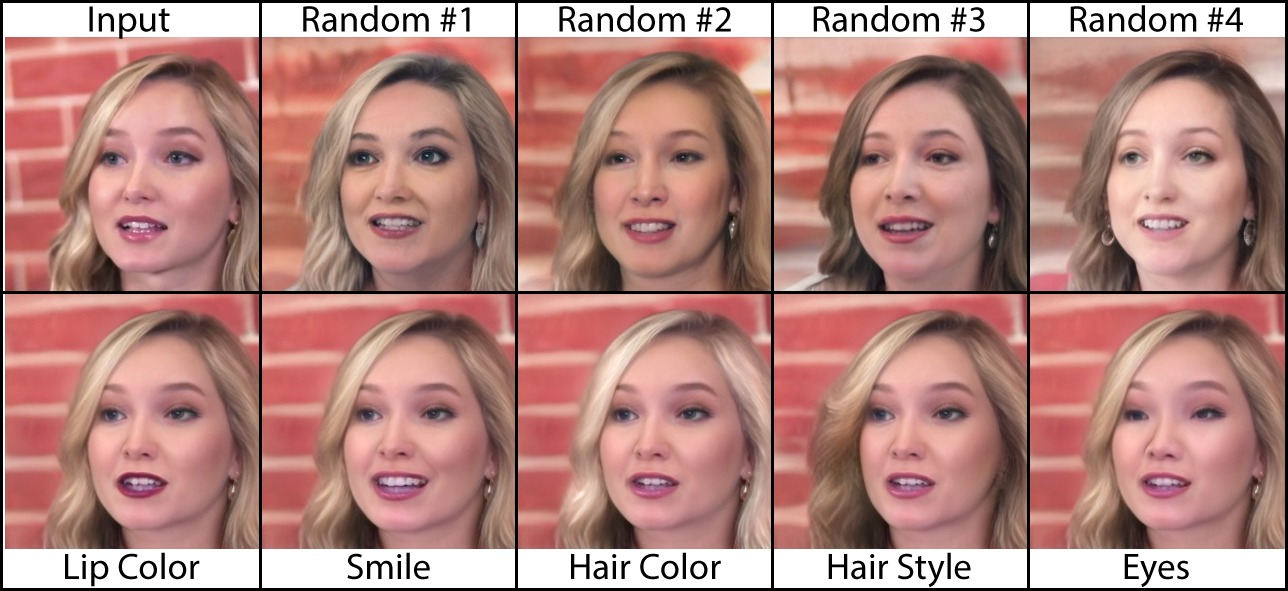}
    \caption{\textit{Appearance editing results for videos. The first row shows the random editings of the input video. The second row shows disentangled editing in StyleSpace~\cite{stylespace}: editing lip color, adding smile, editing hair color and style, editing eye shapes. Best viewed zoomed in. Please watch our supplementary video \#6 for details.}}
    \label{fig:random}
\end{figure}

Another benefit of our work is the convenient face appearance manipulation.
Similar to pose editing, temporal consistency does not need to be considered thanks to our video encoding network.
In the first row of Fig.~\ref{fig:random}, we add a random vector $\delta \mathbf{w}$ to the latent vectors of a face video, resulting in random faces performing the same face motion.
A typical use case of this is protecting the identity in the video by perturbing the look of the person.
Following the previous discussion in StyleSpace~\cite{stylespace}, we can also manipulate individual attributes of the video by adjusting specific dimensions in the StyleSpace.
In the second row of Fig.~\ref{fig:random}, we demonstrate the results of editing individual attributes in the video including lip color, smile, hair color, hairstyle and eye shape.

By learning a fixed direction in the latent space, we can also manipulate the face video with specific semantic meaning.
One example is to edit the age of the face in the video.
For the implementation details of our age editing network, please refer to the supplementary material.
In Fig.~\ref{fig:age}, we demonstrate the results of our latent space age editing for face videos.
Note that wrinkles are learned without explicit efforts to model.
On the right part of Fig.~\ref{fig:age}, we provide age editing results generated by single image method IDInvert~\cite{idgi}.
Since the latent codes for video frames have a significantly different distribution from StyleGAN-like images, the pre-trained latent direction of IDInvert produces wrong attributes like glasses (first three examples) and color bias (last example).
%Note that limited by the age detector supervision, our trained latent space direction cannot change some attributes caused by the decreased age, e.g. hair amount and face shape.
%This can be solved by using a better age detector, but we leave it as a future improvement since it is out of the scope of this paper.

\begin{figure}[t]
    \centering
    \includegraphics[width=0.48\textwidth]{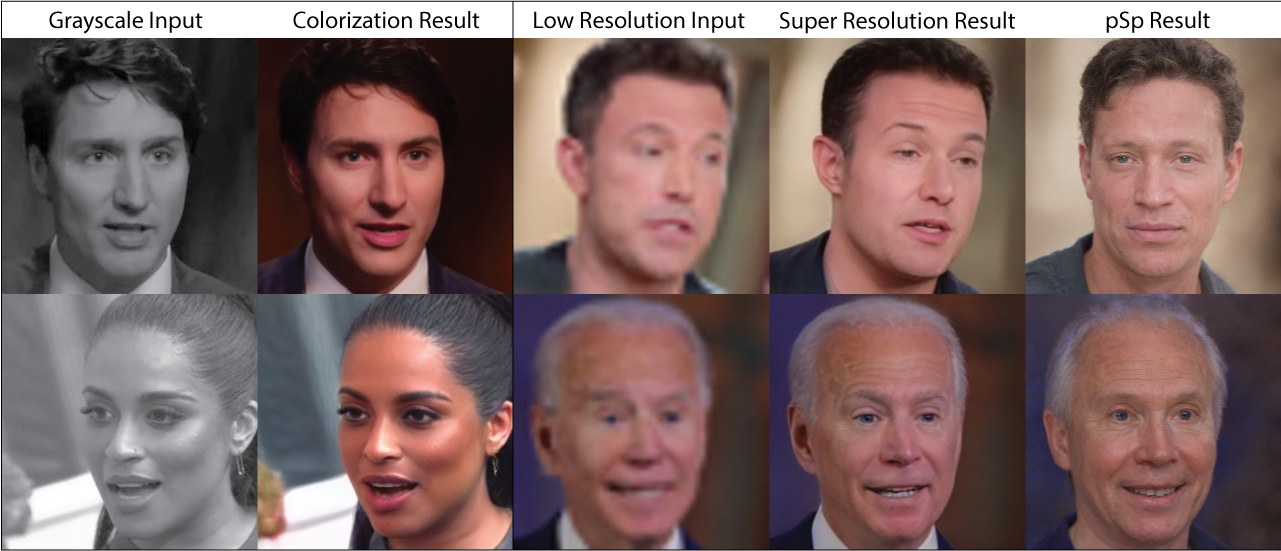}
    \caption{\textit{Colorization and super-resolution of videos using our framework. On the left, by changing the input to grayscale, our network can be trained to encode a single channel video and use StyleGAN to reconstruct its color. On the right, by training our network using downsampled images, our model can also be used as a video super-resolution framework.}}
    \label{fig:color}
\end{figure}

\subsection{Colorization and Super-Resolution}

%Apart from high-level face editing in the video, our face video encoder network also enables common image-domain processing as an image-to-image translation task.
%We demonstrate the ability of video colorization and super-resolution.
%Due to the limited space, please refer to our supplementary material and video \#7 for details.

Apart from face editing in the video, our face video encoder network also enables common image-domain processing as an image-to-image translation task.
Colorization is an operation to recover colored video from the grayscale input video.
Super-resolution is to increase the resolution of the input video while reconstruct the high frequency visual details.
For colorization task, we simply train the video encoding network with grayscale frames.
For super-resolution task, we bicubic downsample the input video then upsampling 4x to its original frame size.
The training procedure of these tasks are the same as described in Sec.~\ref{subsec:training}.
With these simple modifications, our model can be adapted to applications like reconstructing color videos (first row) and improve the visual quality of low-resolution videos (second row), as shown in Fig.~\ref{fig:color}.
The pre-trained super-resolution model of pSp~\cite{psp} yield wrong head pose and identity in videos.

\section{Conclusion}\label{sec:conclusion}
We propose a novel network to encode videos into the StyleGAN latent space.
By reusing the high resolution part of the latent vector, our network enforces temporal consistency with minimum additional computation compared to single frame encoding.
Thanks to our efficient framework, our method achieves real-time performance and is the best compared to existing GAN inversion works.
We introduce facial landmark and 3D face mesh to implicitly guide our network to learn high-level face semantics, handling arbitrary face pose and expression and high fidelity reconstruction of the input video.
We demonstrate the ability to manipulate face videos in the latent space using our model, including pose/expression editing, appearance editing, colorization and super-resolution.
The limitation of our work is that subtle details in the video, like eye gaze and hair dynamics, are not reconstructed well due to the limited expressing power of StyleGAN. 
We believe that the visual quality can be improved using more powerful GANs and loss functions in future works.

%%%%%%%%% REFERENCES
{\small
\bibliographystyle{ieee_fullname}
\bibliography{egbib}
}

\pagebreak
\setcounter{section}{0}
\setcounter{equation}{0}
\setcounter{figure}{0}
\setcounter{table}{0}
\setcounter{page}{1}
\renewcommand\thetable{\alph{table}}
\renewcommand\thesection{\Alph{section}}
\renewcommand\thefigure{\alph{figure}}

%%%%%%%%% TITLE - PLEASE UPDATE
%\begin{widetext}
\begin{center}
\textbf{\large Supplementary Material}
\end{center}
%\end{widetext}

\makeatletter
\renewcommand{\theequation}{S\arabic{equation}}
\renewcommand{\thefigure}{S\arabic{figure}}

\section{Introduction}
This supplementary material is organized as follows.
In Sec.~\ref{sec:age_edit}, we discuss the details of our age editing approach mentioned in the Sec. 4.5 of the main paper.
In Sec.~\ref{sec:addition}, beside Fig. 4 in the main paper, we provide additional qualitative comparison of GAN inversion in videos.
All the examples in this supplementary material can be viewed in dynamics in supplementary video \#3.

\section{Age Editing}\label{sec:age_edit}
\begin{figure}[t]
    \centering
    \includegraphics[width=0.48\textwidth]{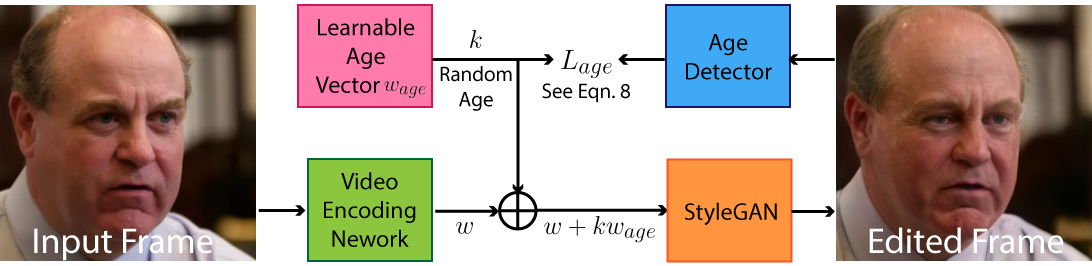}
    \caption{\textit{The structure of our age editing network.}}\label{fig:age-net}
\end{figure}

Our age editing approach is shown in Fig.~\ref{fig:age-net}.
We seek to learn a fixed direction $\mathbf{w}_{age}$ in the latent space, so that the reconstructed face from $\mathbf{w}+k\mathbf{w}_{age}$ is $k$ years younger/older than the input face.
To learn $\mathbf{w}_{age}$, we randomly sample a target age $k$ and an input frame $I$.
Using the pretrained StyleGAN $G$, we obtain the edited frame denoted as $\widehat{I}$:
\begin{equation}
    \widehat{I}=G(E(I)+k\mathbf{w}_{age}).
\end{equation}
We use the pretrained age detector FaceLib~\cite{facelib} to detect the age of the face in the input frame and edited frame.
The training loss aims to make the age difference close to $k$:
\begin{equation}\label{eqn:age}
    L_{age}=\left \|k-(D(I)-D(\widehat{I})) \right \|_2 +\lambda_{f}L_{f}+\lambda_{v} L_{v},
\end{equation}
where $D$ represent the age detected from the frame.
In Eqn.~\ref{eqn:age}, to constrain the other attributes of the face, like pose and expression, we also impose the sparse facial landmark loss $L_f$ and dense 3D loss $L_v$ introduced in Sec.~3 in the main paper.

\section{Additional Comparison}\label{sec:addition}
In addition to Fig.~4 in the main paper, we provide more qualitative comparison with pSp~\cite{psp}, e4e~\cite{e4e}, ReStyle~\cite{restyle}, IDInvert~\cite{idgi} and Image2StyleGAN~\cite{image2stylegan} in Fig.~\ref{fig:inv1}. 
These examples cover various challenging scenarios for high-fidelity GAN inversion:
\textbf{I) Normal head pose}: Examples (1), (2), (3), (10), (14), (19), (21), (24) and (26) are talking videos with normal head poses.
These type of videos is relatively easy for existing single image GAN inversion methods. 
\textbf{II) Large expression}: Examples (4), (5), (9), (13), (15), (16), (20), (22), (25) and (27) contains large expression like closed eyes. 
These videos require high accuracy in GAN inversion since human is sensitive to unnatural face movements in the video. 
\textbf{III) Large head pose}: Examples (6), (7), (8), (11), (12), (17), (18) and (23) involve with large head pose in the image space. 
Existing single image GAN inversion methods could not handle this case in general, since they assume center-aligned face in the input image and are not aware of the actual pose and expression of the face. 

According to the qualitative results, we have the following observations:

\textbf{1)}It is worth noting that existing GAN inversion methods like pSp, e4e and their subsequent work ReStyle cannot guarantee the accurate identity and expression reconstruction in videos, even for simple cases like examples (1), (2) and (3).

\textbf{2)}IDInvert also fails catastrophically in many videos, e.g., examples (7), (12), (15) and (20). 
Some of the failure patterns (e.g. (15) and (20) are highly similar, indicating that many input video frames reside outside the domain of FFHQ images and causing the saturation of the pre-trained GAN inversion network.
This further support our claim that existing single-image GAN inversion approaches cannot handle videos in a straightforward way.

\textbf{3)}Optimization based method Image2StyleGAN~\cite{image2stylegan} sometimes reconstruct the pose and expression accurately, like examples (2), (5), (7) and (12), but it requires more than 5min (1500 iterations) per frame and the visual quality is still inferior to our method. 
Although it is possible that their result is able to converge to the input frame exactly, the extremely long processing time is prohibitive to video applications.

\textbf{4)}With high-level facial landmark loss and face mesh loss, our model is able to handle arbitrary pose and expression, even in large pose and expression cases like examples (4), (7), (12) , (17) and (18).

To sum up, our method is the only method that can handle video GAN inversion, and is the only one achieving real-time performance among comparison methods.

\begin{figure*}[h]
\includegraphics[width=\textwidth]{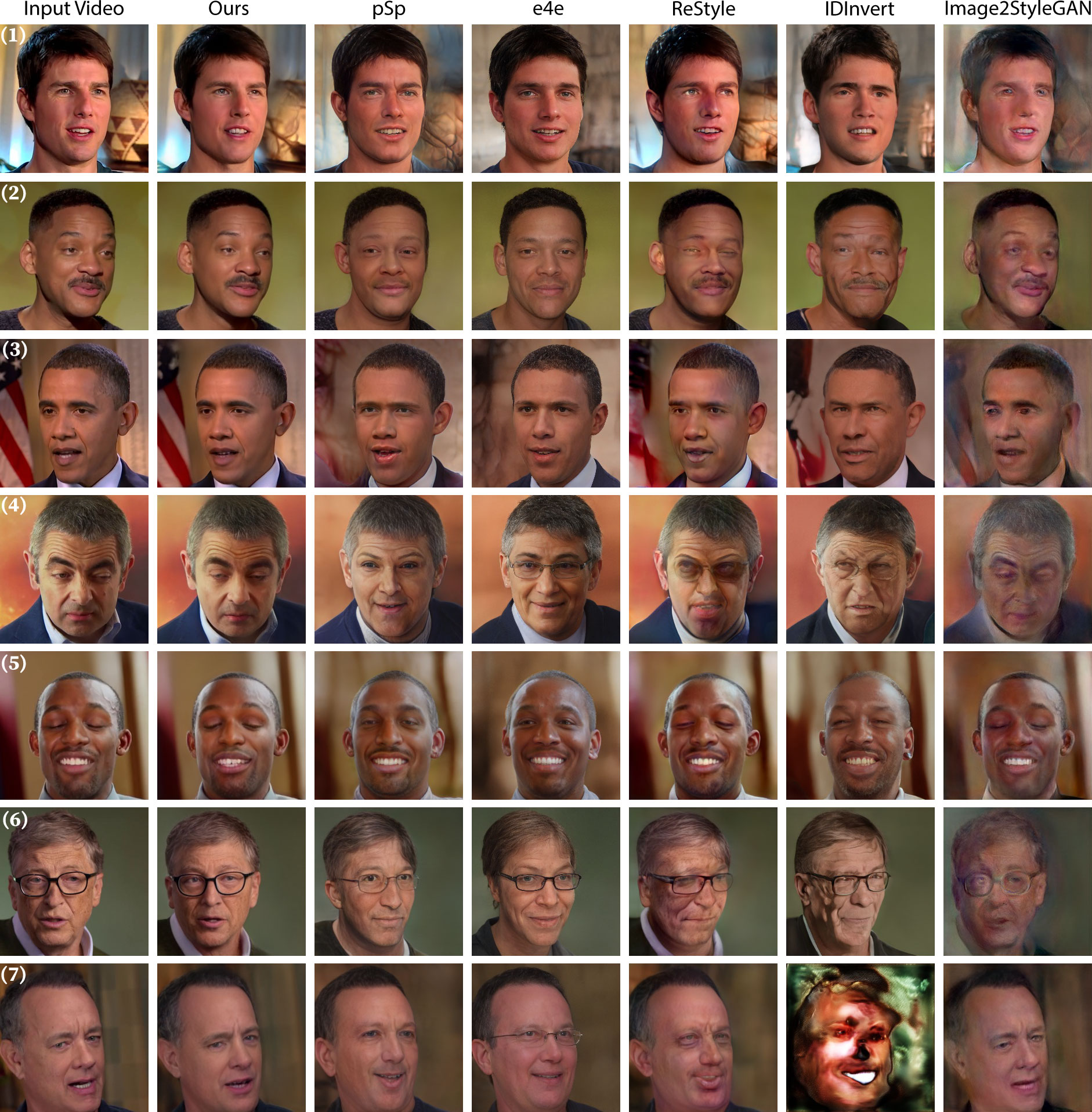}
\caption{\textit{Additional qualitative comparison to Fig.~4 in the main paper. Examples cover various challenging scenarios for high-fidelity GAN inversion.
Best viewed zoomed in. Please watch our supplementary video \#3 for details.}}
\label{fig:inv1}
%\vspace{-15pt}
\end{figure*}

\begin{figure*}[t]
\ContinuedFloat
\captionsetup{list=off,format=cont}
\includegraphics[width=0.9\textwidth]{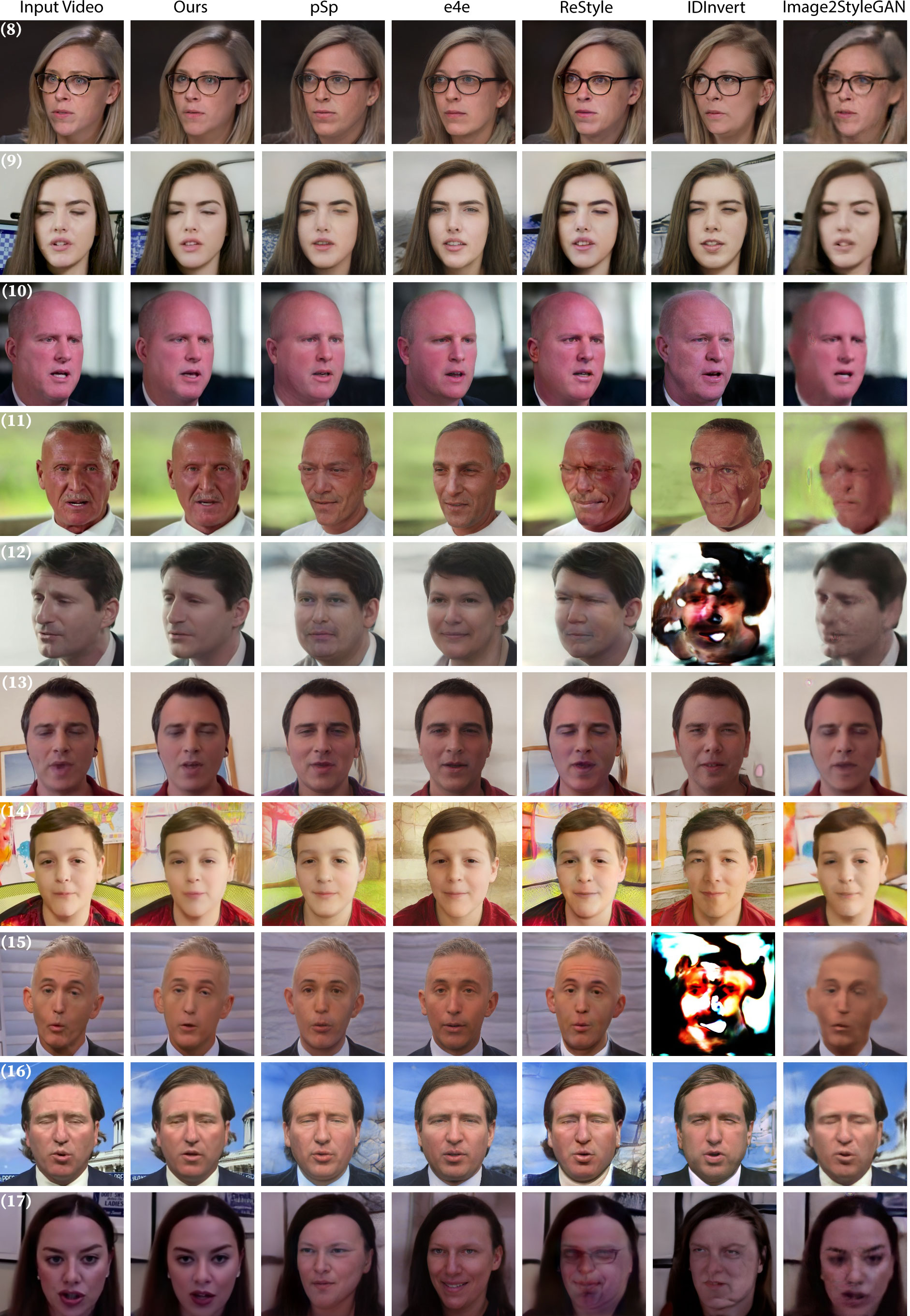}
\caption{\textit{Additional qualitative comparison to Fig.~4 in the main paper. Examples cover various challenging scenarios for high-fidelity GAN inversion.
Best viewed zoomed in. Please watch our supplementary video \#3 for details.}}
\label{fig:inv2}
\end{figure*}

\begin{figure*}[t]
\ContinuedFloat
\captionsetup{list=off,format=cont}
\includegraphics[width=0.9\textwidth]{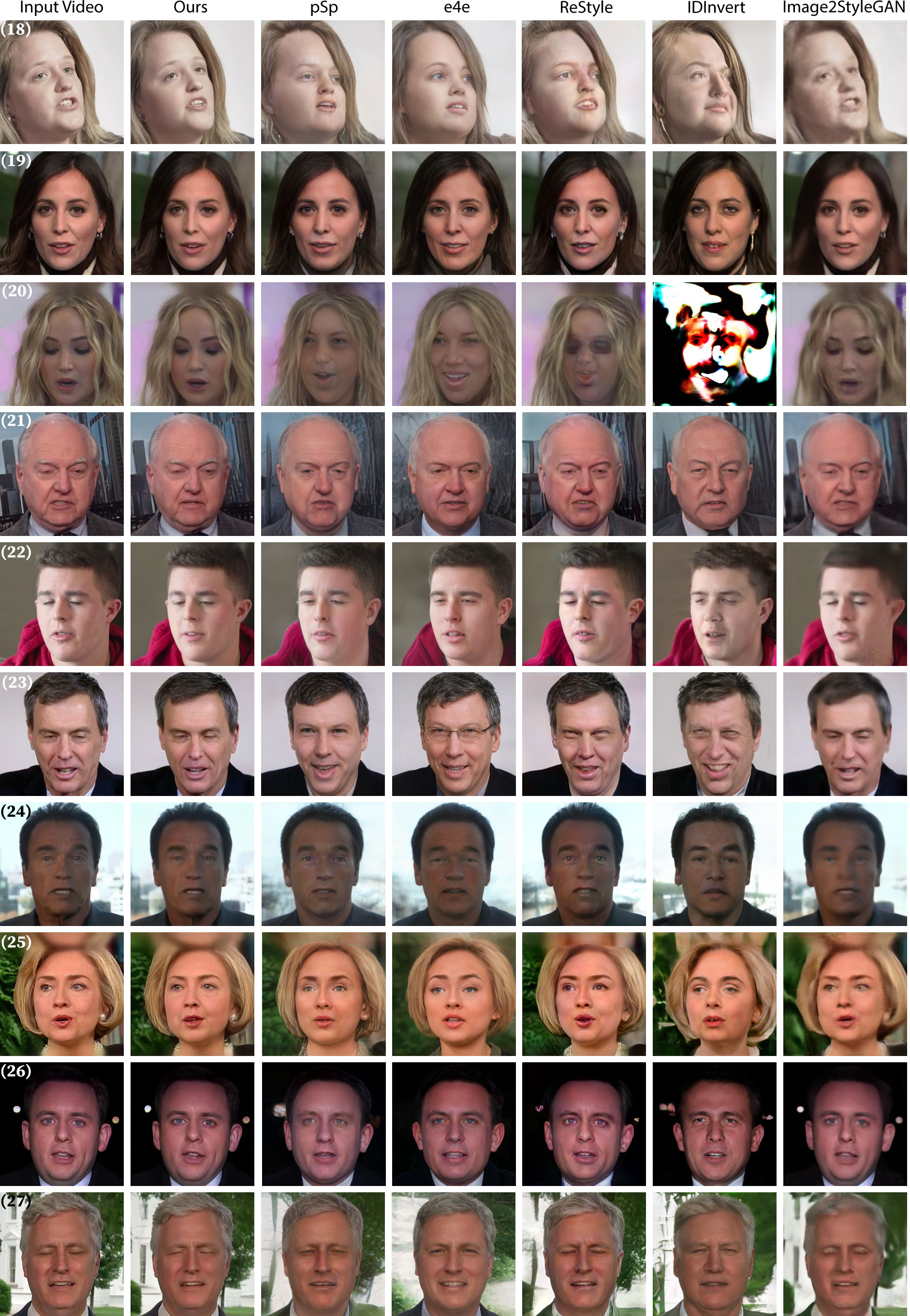}
\caption{\textit{Additional qualitative comparison to Fig.~4 in the main paper. Examples cover various challenging scenarios for high-fidelity GAN inversion.
Best viewed zoomed in. Please watch our supplementary video \#3 for details.}}
\label{fig:inv3}
\end{figure*}

%\FloatBarrier
%%%%%%%%% REFERENCES
%{\small
%\bibliographystyle{ieee_fullname}
%\bibliography{egbib}
%}
%\end{document}

\end{document}